\documentclass[letterpaper, 10 pt, conference]{ieeeconf}  

\IEEEoverridecommandlockouts                              

\usepackage{cite}
\usepackage{amsmath,amssymb,amsfonts}
\usepackage{algorithmic}
\usepackage{graphicx}
\usepackage{textcomp}
\usepackage{xcolor}
\def\BibTeX{{\rm B\kern-.05em{\sc i\kern-.025em b}\kern-.08em
    T\kern-.1667em\lower.7ex\hbox{E}\kern-.125emX}}

\usepackage[breaklinks=true,bookmarks=false]{hyperref}

\usepackage{subcaption}

\title{\LARGE \bf
Off-the-shelf bin picking workcell with visual pose estimation: \\
A case study on the world robot summit 2018 kitting task
}

\author{Frederik Hagelskjær$^{1}$, Kasper Høj Lorenzen$^{1}$ and Dirk Kraft$^{1}$
\thanks{This project was funded in part by Innovation Fund Denmark through the project MADE FAST, in part by the SDU I4.0-Lab.}
\thanks{
All authors are with SDU Robotics, Mærsk Mc-Kinney Møller Institute, University of Southern Denmark, 5230 Odense M, Denmark
        {\tt \{frhag,kalor,kraft\}@mmmi.sdu.dk}}%
}

\begin{document}

\maketitle
\thispagestyle{empty}
\pagestyle{empty}

\begin{abstract}
The World Robot Summit 2018 Assembly Challenge included four different tasks. The kitting task, which required bin-picking, was the task in which the fewest points were obtained. 
However, bin-picking is a vital skill that can significantly increase the flexibility of robotic set-ups, and is, therefore, an important research field.
In recent years advancements have been made in sensor technology and pose estimation algorithms. These advancements allow for better performance when performing visual pose estimation. 

This paper shows that by utilizing new vision sensors and pose estimation algorithms pose estimation in bins can be performed successfully. We also implement a workcell for bin picking along with a force based grasping approach to perform the complete bin picking.
Our set-up is tested on the World Robot Summit 2018 Assembly Challenge and successfully obtains a higher score compared with all teams at the competition.
This demonstrate that current technology can perform bin-picking at a much higher level compared with previous results.
\end{abstract}


\section{Introduction}

Bin-picking is fundamental task for industrial robotics. It allows for object feeding without a need for fixtures or manual insertion. This makes robotic solutions much more flexible and allows for adaptive production. 

A simple approach for bin picking, is visual pose estimation. 
However, the set-up of a pose estimation solution is a difficult task. It can be very time consuming to fine tune parameters to obtain adequate performance \cite{hagelskjaer2018does}. 
Especially for industrial objects with shiny metallic surfaces obtaining precise pose estimation can be very difficult. When integrating grasping this is further complicated \cite{hagelskjaer2019combined}.
%
%
%
%
%
This was especially demonstrated in the World Robot Summit Assembly Challenge in 2018 (WRSAC18) Kitting Task \cite{yokokohji2019assembly}. 
WRSAC18 had robotic workcells from all over the world competing in industrial robotics. The challenge consisted of four different tasks in which a robot should manipulate industrial objects. The competitors showed impressive results with state-of-the-art methods.

\begin{figure}[t]
    \begin{center}
    \begin{subfigure}[t]{.23\textwidth}
      \centering
      \includegraphics[trim=0 1100 0 1150,clip,width=0.99\linewidth]{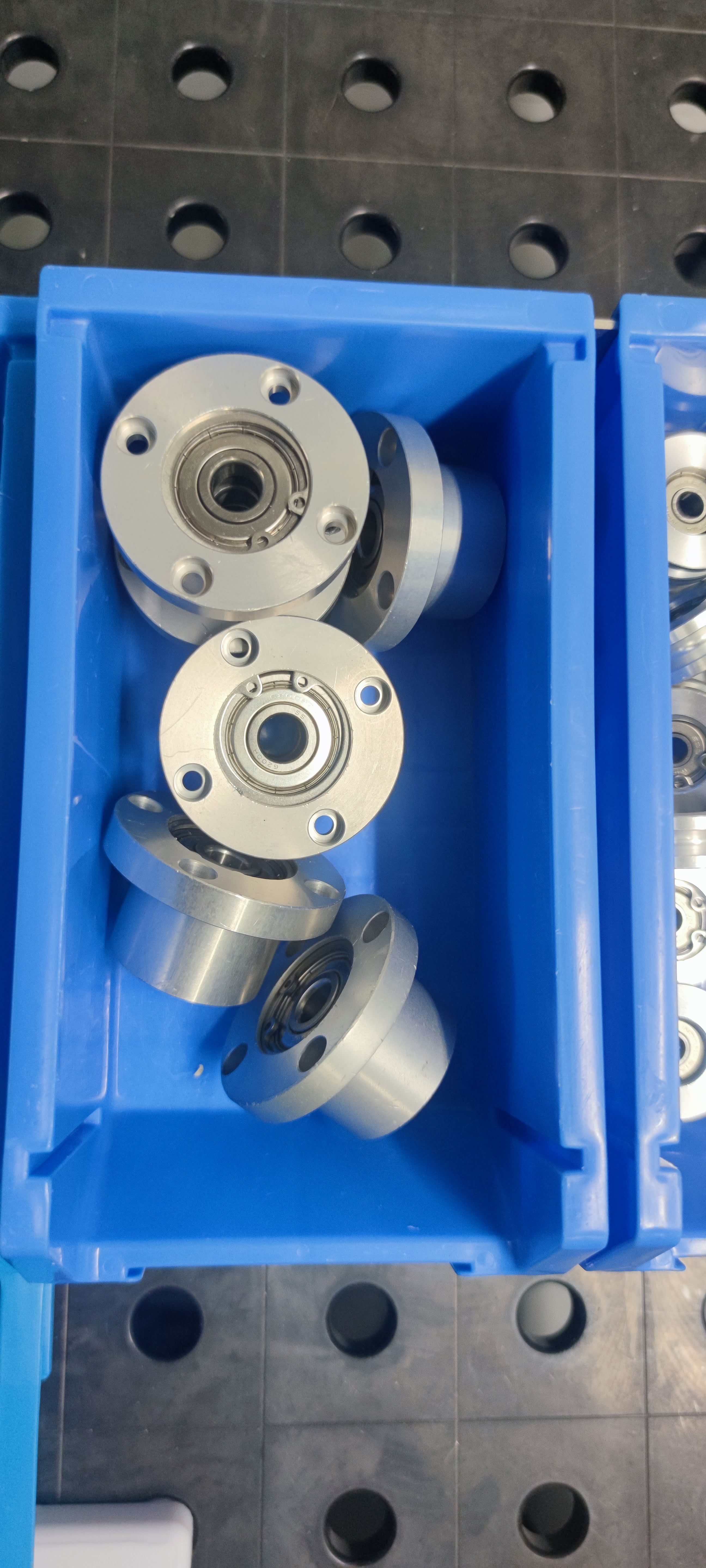}
      \caption{Objects in bin.}
      \label{fig:grasp:01}
    \end{subfigure}%
    ~
    \begin{subfigure}[t]{.23\textwidth}
      \centering
      \includegraphics[trim=900 500 1300 450,clip,width=0.99\linewidth]{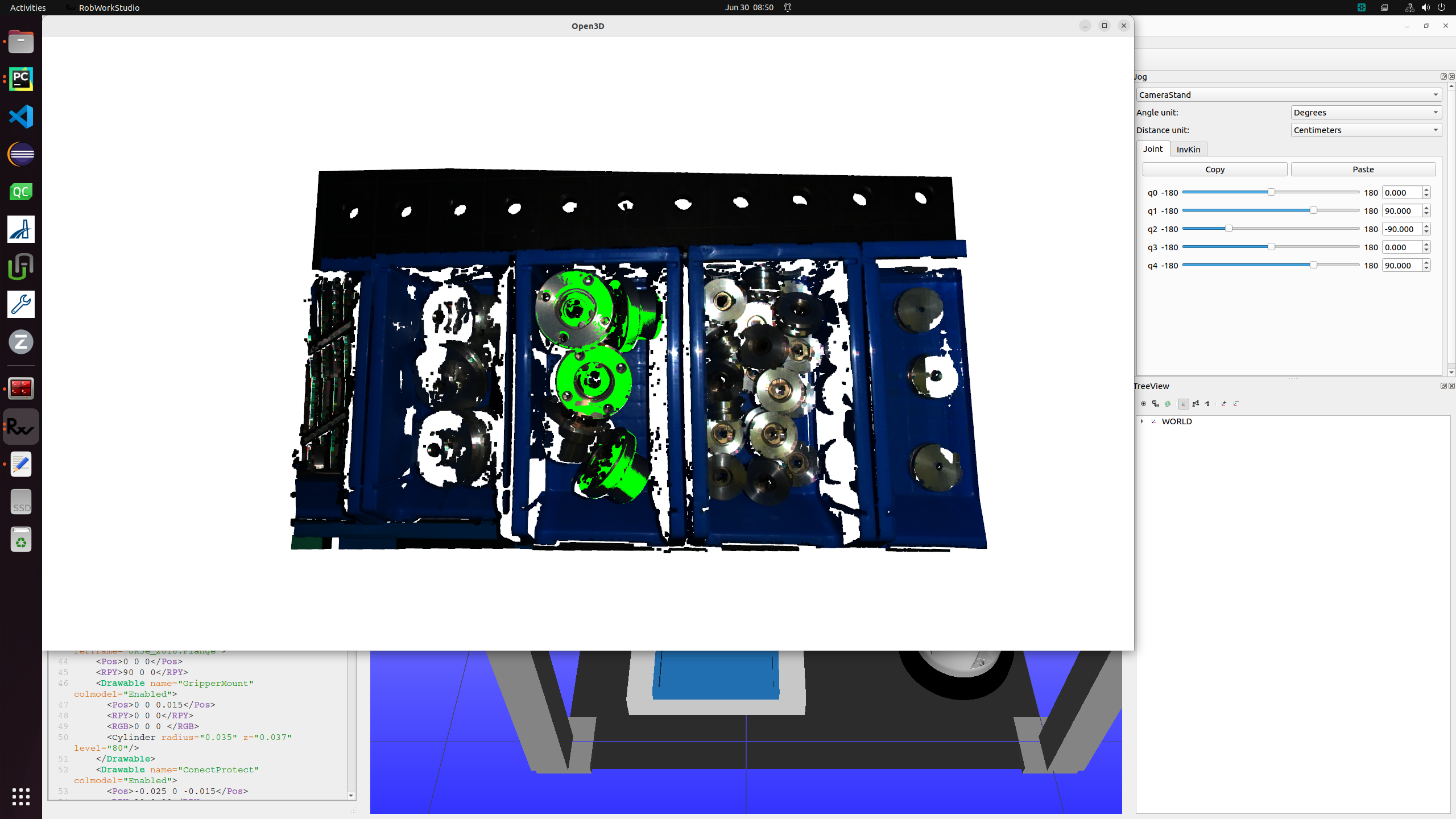}
      \caption{Pose Estimation.}
      \label{fig:grasp:02}
    \end{subfigure}%

    \begin{subfigure}[t]{.23\textwidth}
      \centering
      \includegraphics[trim=1350 300 600 300,clip,width=0.99\linewidth]{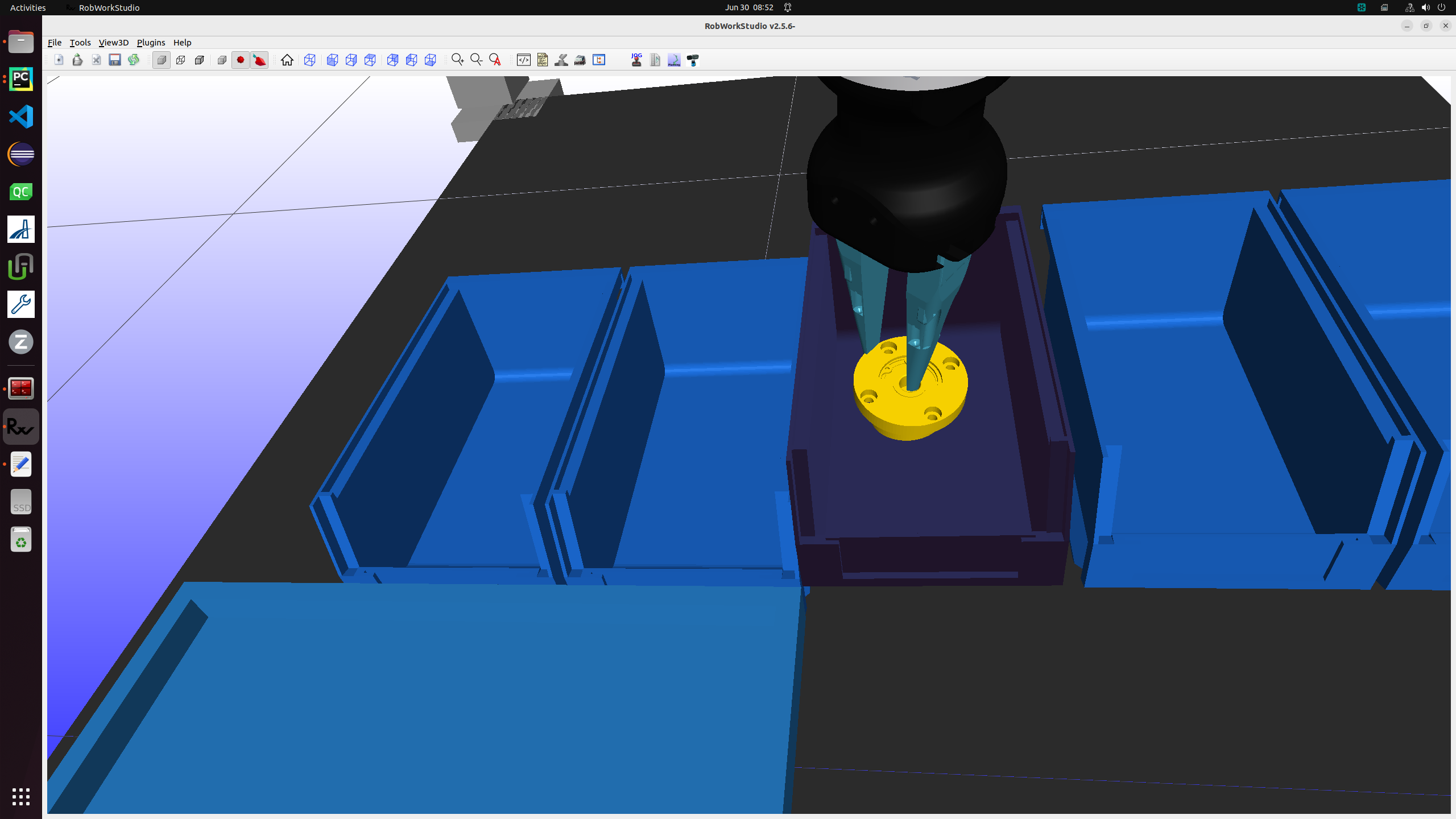}
      \caption{Grasp pose planning in simulation.}
      \label{fig:grasp:03}
    \end{subfigure}%
    ~
    \begin{subfigure}[t]{.23\textwidth}
      \centering
      \includegraphics[trim=0 1000 0 1200,clip,width=0.99\linewidth]{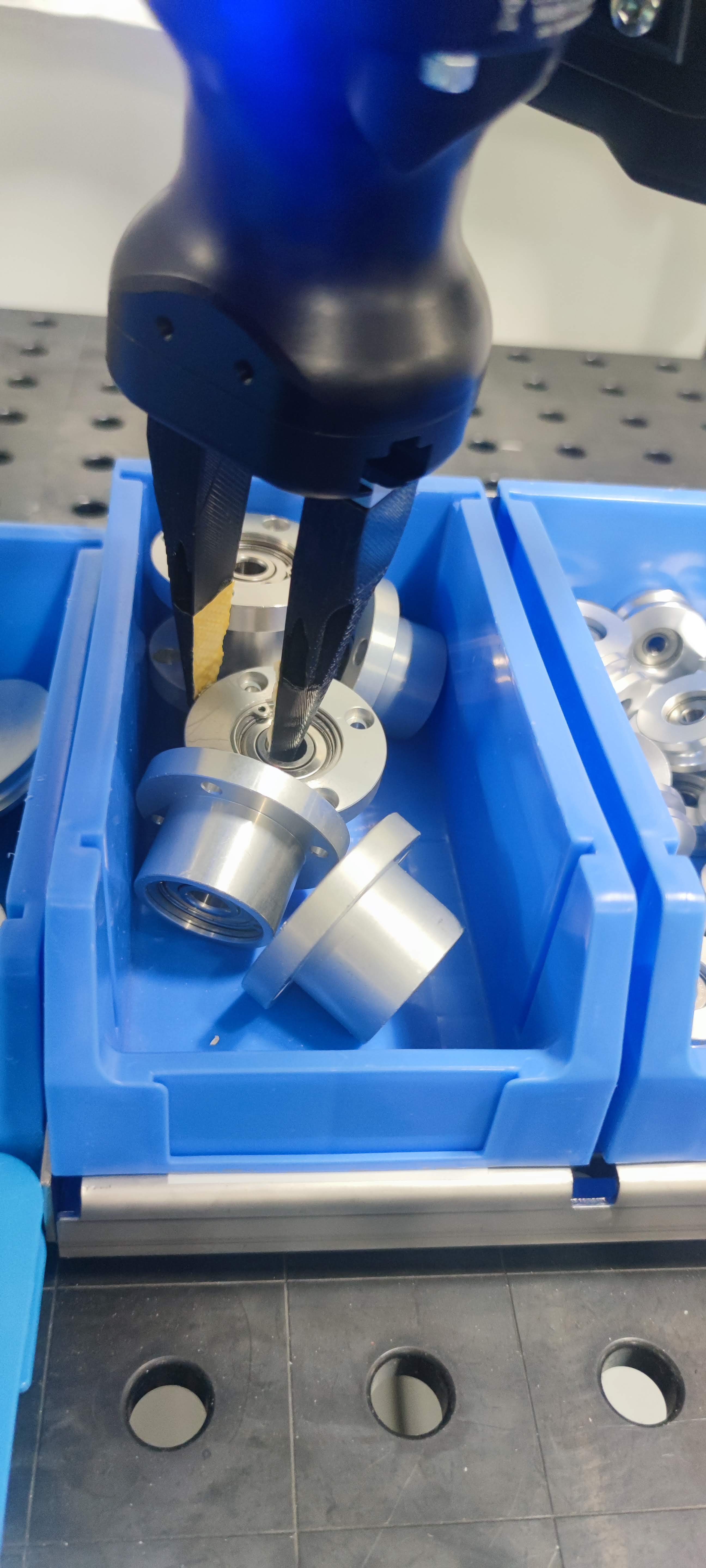}
      \caption{Grasp executed in real world.}
      \label{fig:grasp:04}
    \end{subfigure}%
    
   \caption{The different steps in the bin picking. Initially the object is placed in the bin. From the point cloud the poses are estimated. The best grasp pose is found in simulation, and finally the object is grasped.}
   \label{fig:grasp}
     \vspace{-6mm}
\end{center}
\end{figure}

However, the bin-picking task in particular, was very difficult for all teams. In the task fifteen different object varying in size and shape were to be grasped. The objects were placed in homogenous bins, and were to be placed in a kitting tray in specific positions. During competition ten selected objects were to be placed on kitting trays, repeated three times, with 20 minutes for completion, giving 40 seconds per object.
Two points were earned per object, with an additional thirty points for a completed board, and with three repetitions this meant that 150 points could potentially be achieved \cite{wrsrules}.
However, the highest-scoring team only obtained 20 points. This meant that the best-performing team, accomplished ten of the thirty bin-picks in twenty minutes. 
Additional analysis of all teams, shows the object success rate of the objects, with the highest being 28.1~\% and the lowest being 1.6~\%. Analysis of the time spent on each task by the team showed that on average 21~\% of the time was spent on this task. The lacking performance was thus not simply a result of not prioritizing the task \cite{von2020robots}.
Additionally, the task was removed for the subsequent challenge in 2020 \cite{yokokohji2022world}.


In this paper we introduce an off-the-shelf bin picking workcell. 
The developed workcell is tested on the WRSAC18 kitting task for the large objects. These were the objects with the lowest score during the challenge \cite{yokokohji2019assembly}. 
The workcell uses a modern depth sensor, the Zivid2 to obtain point clouds. We employ a modified version of a state-of-the-art pose estimation algorithm \cite{hagelskjaer2022parapose}.
A robust bin picking procedure then grasps the objects, and they are finally placed in the kitting tray.
%

Our developed workcell demonstrate good results for bin-picking. The workcell is able to successfully complete the kitting task for all large objects. The kitting lasts 28 seconds on average, and with an average of 40 seconds available per object, completing the full kitting is possible.

%
We believe that the results of this workcell demonstrate the development in bin-picking in recent years compared with the results at WRSAC18.




In this paper the following main contributions are presented.
\begin{itemize}
    \item Adapted state-of-the-art algorithm for colorless pose estimation 
    \item Implemented collision prediction grasp strategy for bin picking
    \item A flexible workcell for automatic set-up of bin picking
    \item State-of-the-art results for the WRSAC18 kitting task for large objects
\end{itemize}

The remaining paper is structured as follows: We first review related papers in Sec.~\ref{related}. In Sec.~\ref{met}, our developed method is explained and the different contributions are elaborated. In Sec.~\ref{exp}, the experiments are performed, and the performance is verified. Finally, in Sec.~\ref{con}, a conclusion is given to the paper, and further work is discussed.

\section{Related Work}
\label{related}

As bin picking is an important topic in creating flexible robotic set-ups, several different approaches have been developed. Solutions range from data driven \cite{kozak2021data}, to based on simulations \cite{schyja2015realistic} and digital twin solutions \cite{tipary2021generic}. Cameras are generally a part of the solution \cite{kristensen2001bin, schyja2012modular}.









At WRSAC18 several different bin picking solutions were presented. 
%
%
%
SDU Robotics was the team that obtained the overall highest score during the competition \cite{schlette2020towards}. The solution presented by SDU Robotics consisted of two different solutions for small and large objects. For the small object a scooping mechanism was developed \cite{mathiesen2019vision}. This allowed for mechanical singulation by shaking the object after scooping. A second robot then grasps the object from the scoop. 
However, this approach was not suitable for the larger objects, and a suction approach was created instead. After the grasping the object position is unknown in the hand and to perform the kitting the pose needs to be found. This was performed by placing the object on a re-grasping table, calculating the pose using template matching \cite{hagelskjaer2019using} and grasping the object with a fingered gripper \cite{schlette2020towards}.

The team which obtained the second highest overall score was JAKS \cite{tajima2020robust}. JAKS used a combined 2D/3D vision system for pose estimation combined with pre-registered grasp poses. Using the robot and bin model they check the grasps for collision. Our method also employs collision check for grasping, however we also test for collision with the point cloud data using the sensors, and we allow for small collisions with the fingers. 
JAKS pose estimation system is based on LINEMOD \cite{hinterstoisser2012model}, a classical pose estimation method based on template matching. The pose estimation system does not work for the smaller objects and a Hough \cite{yuen1990comparative} based search is used instead. Tactile sensing is used in the finger for detecting grasps, and the encoders in the gripper are used to verify the correct grasp. We employ the same strategy of using the encoder to verify that an object is grasped, however the gripper and not tactile sensors are used to detect grasps. 


The team which obtained the highest score in the kitting task was Robotic Materials \cite{yokokohji2019assembly}. Their method relied on a gripper with an inbuilt 3D sensor \cite{correll2021systems, watson2020autonomous}. Their gripper is shaped with long narrow fingers which allow for moving into the bin. We employ the same design for our fingers. 
They employ grasp primitives, which could potentially lead to grasping of unknown objects. 
In our application the grasp poses are defined beforehand to ensure the object is in a known pose. 
The grasping strategy of Robotic Materials allows for approaching in inclined angles, which allow for more grasp poses. Our method utilize the same approach to allow us to always have possible grasp poses.
For some of the small objects, they used jigs were to reorient the objects \cite{von2020robots}. 
In this paper we focus on the larger objects, and have not utilized any jigs.


The team with the second highest score for the kitting challenge was Cambridge Robotics \cite{hughes2020flexible}. The team used an alternative gripper with adhesive pad. The pad is moved towards the object until contact is detected. The adhesive surface holds the object while the robot moves from the bin to the kitting tray. A mechanism then releases the object by pushing it off the adhesive. The objects are detected using a neural network trained on 15,200 images. The paper shows results for only the small objects, as the adhesive would be less effective for heavier objects. The approach obtains good results, with some errors when the objects were smaller than the release mechanism, and thus stayed attached to the finger.

In a summary of the competition \cite{von2020robots} it was speculated that the success of SDU Robotics was middle ground top-of-the-line industrial solutions and custom made solutions. 
Our implementation for the kitting task follows the same approach of a middle ground. All hardware for the task, i.e., robot, gripper and 3D sensor, are top-of-the-line industrial products. The software used is custom designed systems created for bin-picking. Additionally, the fingers are custom made for the bin-picking task.

\section{Method}
\label{met}

\begin{figure}[ht]
\begin{center}
   \includegraphics[width=0.65\linewidth]{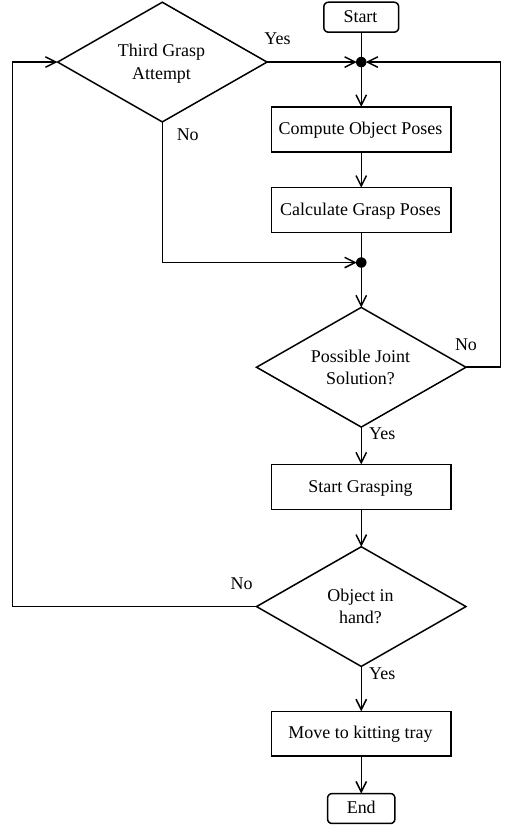}
   \caption{The overall workflow for a bin picking operation.}
   \label{fig:full_workflow}
    \vspace{-6mm}
\end{center}
\end{figure}

The presented method is a pipeline for off-the-shelf bin picking. The method consists of both an offline set-up phase and an online run-time phase. The offline set-up phase consists of training the pose estimation algorithm and defining the grasp poses.
During run-time, the robot moves the camera above the bin, object poses are computed, the best grasp pose is calculated and the robot attempts to grasp the object. This process is repeated until an object is grasped. The full pipeline is shown in Fig.~\ref{fig:full_workflow}

In the following section the different parts of the bin picking set-up is elaborated. The workcell is described in Sec.~\ref{met:work}, in Sec.~\ref{met:pos} the pose estimation algorithm is explained, Sec.~\ref{met:grasp} contains the grasp poses, and finally in in Sec.~\ref{met:exe} the grasp executor is detailed.

\subsection{The Workcell} \label{met:work}

\begin{figure*}[ht]
    \vspace{1.5mm}
    \begin{center}
    \hfill
    \begin{subfigure}[t]{.43\textwidth}
      \centering
      \includegraphics[trim=1500 0 1500 0,clip,width=0.99\linewidth]{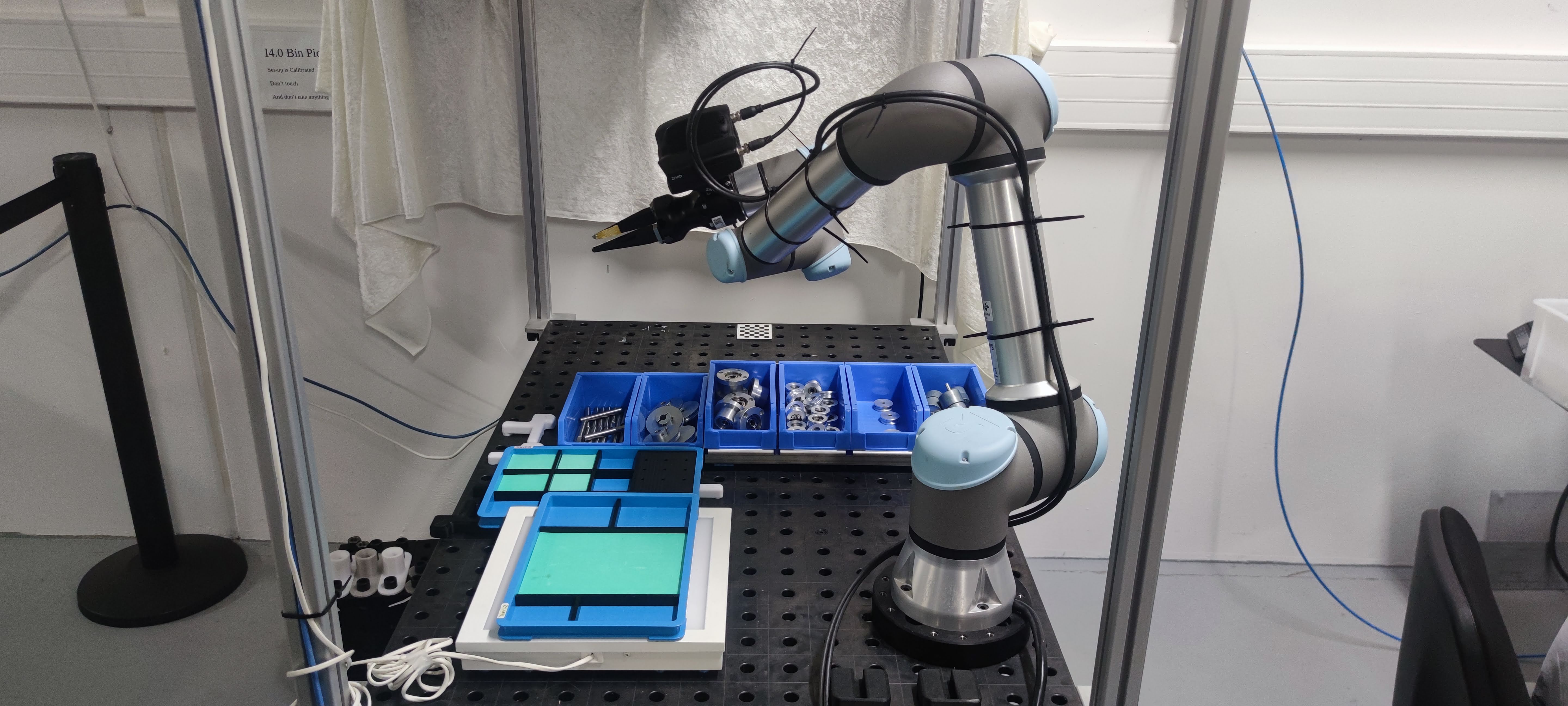}
      \caption{Real workcell.}
      \label{fig:workcell:01}
    \end{subfigure}%
    \hfill
    \begin{subfigure}[t]{.43\textwidth}
      \centering
      \includegraphics[trim=900 300 600 150,clip,width=0.99\linewidth]{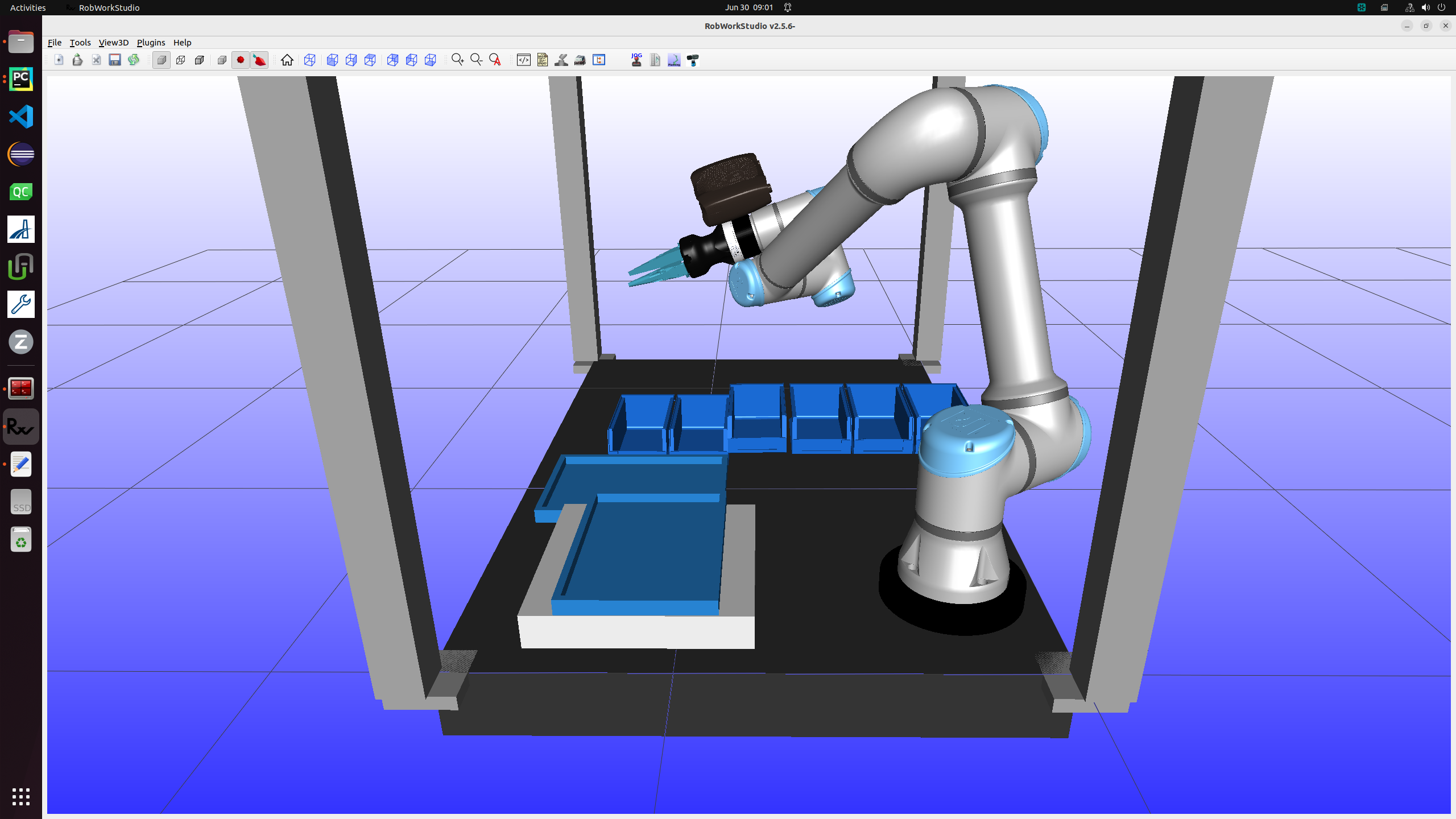}
      \caption{Digital twin.}
      \label{fig:workcell:02}
    \end{subfigure}%
    \hfill
    ~

   \caption{The used workcell to accomplish the bin picking. The real workcell is shown to the left. To the right the Digital Twin is shown. The scene consists of the robot with gripper and camera, the bins (center) and kitting trays (bottom left). All actions are relative to the objects and the bins and trays can, therefore, be moved around in the workcell freely.}
   \label{fig:workcell}
\end{center}
\end{figure*}

The workcell for the bin picking consists of an UR5 robotic arm, a Robotiq HAND-E gripper with 3D printed fingers and a Zivid2 robot-mounted depth sensor. The robot is placed on a table along with the trays and bins. The workcell uses a PC environment with an Intel i9-9900K 3.60GHz CPU and an NVIDIA GeForce RTX 2080 TI GPU.


In the bin-picking task, the objects are placed randomly in the bin. Thus the grasp poses cannot be computed offline. A digital twin of the workcell has been created to avoid collisions when grasping. This allows the system to check all grasps poses for collision before execution. Additionally, by configuring the positions in the digital twin all actions are defined relative to the trays and bins. Thus multiple bins and trays can be placed around the workcell, and the robot can plan collision free solutions.
%
The real workcell and digital twin are shown in Fig.~\ref{fig:workcell:02}. The digital twin is implemented using Robwork \cite{ellekilde2010robwork}.


\subsubsection{Finger Design}

\begin{figure}[ht]
\begin{center}
   \includegraphics[width=0.75\linewidth]{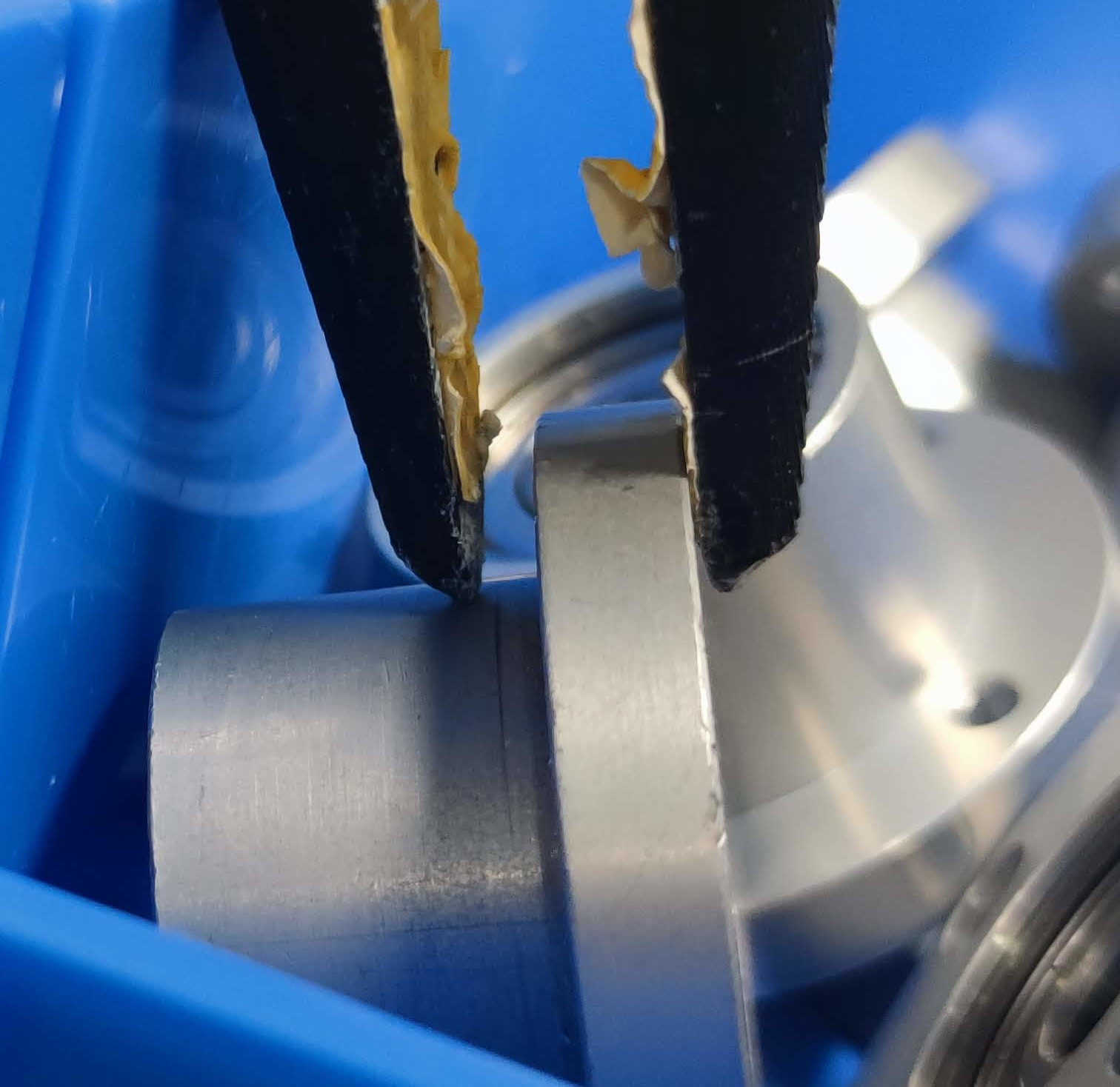}
   \caption{The fingers grasping object type 7 in the bin.}
   \label{fig:fingers}
     \vspace{-6mm}
\end{center}
\end{figure}

The finger design is inspired by the Robotic Materials finger \cite{watson2020autonomous}. The sharp curvature at the fingertip allows for moving into the bin without being blocked by minor collisions. Rubber is applied to the inside of the fingers to improve the friction of the grasp. The fingertips are shown grasping an object in Fig.~\ref{fig:fingers}.


\subsection{Pose Estimation Algorithm} \label{met:pos}
To grasp and correctly place the objects, pose estimation is necessary. To perform the pose estimation, a variant of the ParaPose \cite{hagelskjaer2022parapose} algorithm has been implemented. This algorithm has been selected for several reasons. Firstly the algorithm has state-of-the-art performance on the challenging Occlusion dataset \cite{brachmann2014learning}, secondly, the algorithm has a completely automatic set-up using synthetic data, and thirdly, the algorithm uses 3D point clouds for the pose estimation. 

The 3D input data is important as the provided CAD models often do not contain surface or color information. This generally holds for most CAD models in manufacturing industry. Pose estimation algorithms without color information are thus necessary.
To accommodate this need, the ParaPose algorithm is adapted to not use color information, and the RGB input is removed from the network. 
%


\textbf{Instance Segmentation:} 
As the objects are placed in bins with homogeneous content, the detection task is vastly simplified. Using the digital twin of the workcell, the position of the bin is approximately known. Iterative Closest Point (ICP) is then used to obtain the actual position of the bin. The bin and remaining scene can then be removed from the point cloud, and only points belonging to the object remain. 

As in PointVoteNet \cite{hagelskjaer2020pointvotenet}, we sample anchor-points for the algorithm, which are then used for pose estimation. The task is thus reduced to instance segmentation for each point cloud at the anchor points.

The full pose estimation procedure is as follows:

\begin{itemize}
    \item Refine bin position with ICP
    \item Remove all points outside the bin
    

    \item Generate anchor points
        
    \item For each anchor point, sample point cloud based on radius
    \item Process each point cloud with the network and compute matches
    \item Perform multiple pose estimations for each point cloud and refine with ICP
    \item Sort according to depth check
    \item Non-maximum suppression based on ADDS \cite{hinterstoisser2012model}
\end{itemize}


\textbf{Set-up:} 
We use the same set-up for ParaPose as in the original paper \cite{hagelskjaer2022parapose}. However, the synthetic data is adapted to the bin-picking scenario. The data is generated using the BlenderBin\footnote{\url{https://github.com/hansaskov/BlenderBin/}} pipeline. BlenderBin is an extension of the BlenderProc \cite{denninger2019blenderproc} pipeline with a focus on bin-picking. The objects are placed in homogeneous bins replicating the real world scenario. Thus synthetic data is easily created and the network is trained using this data.

During the network training the domain randomization is automatically learned. As color information has been omitted the RGB noise has been removed from the parameter tuning. While the pose estimation parameters are also optimized automatically in the original approach, we implement heuristic parameters in this method. The heuristic approach was chosen as the workcell remains static. Parameters will thus not change for new objects, and manual tuning is thus feasible.

\subsection{Grasp Poses} \label{met:grasp}
As the objects are placed freely in six degrees of freedom (6DOF), the world relative grasp poses are also freely placed in 6DOF. Thus a grasp pose can easily be invalided either because it cannot be reached, or as a result of collision with the workspace. To ensure that the found objects can be grasped, grasp poses should be defined all around the object. If only a single grasp pose is defined, the object would often not be graspable.
%
Our workcell thus work best with method for automatically creating grasp poses. These approaches could be e.g., simulation such as GraspIt \cite{miller2004graspit} or geometric primitives \cite{ti2023geometric}.

\textbf{Cylindrical Grasp Poses:} As all the objects in this task are cylindrical in shape we have created a grasp pose generator based on this. A single good grasp pose can thus be used to generate a set of grasps poses around the object. By rotating 360 degrees around the object along with a single 180 degree around the hand (taking the symmetry of the parallel gripper into account) a full set of grasp poses can be created. 
Using this method, the robot can create effective grasp poses from any cylindrical object. Compared with generating grasp poses from simulation this allows for consistent grasps independent of the object pose.


\textbf{Computing Grasp Solutions:} When the pose estimation has been performed, the set of all grasp poses, $_{base}T_{tcp}$, can be computed using Eq.~\ref{eq:grasppose}. Were $_{base}T_{obj}$ is the set of objects poses and $_{obj}T_{tcp}$ is the set of grasp poses.

\begin{equation} \label{eq:grasppose}
 \ _{base}T_{tcp} = \ _{base}T^{n}_{obj}  \ _{obj}T^{m}_{tcp}
\end{equation}

This results in $n \times m$ solutions for possible grasps in TCP space. Using the analytical solver \cite{ellekilde2010robwork} this returns $8 \times n \times m$ possible joint configurations. The goal is then to find the grasp pose with the shortest distance in joint space. As the joints can generally move in parallel, only the largest joint distance is relevant. The best grasp pose can thus be calculated as in Eq.~\ref{eq:minmax}.


\begin{equation} \label{eq:minmax}
\min_{\forall s \in S} ||j_{s} - \hat{j}||_{0}
\end{equation}

Where $S$ is the space of all grasping solutions and $\hat{j}$ is the current joint configuration. For each successive joint configuration path planning is performed. If a collision free path cannot be found, the solution is discarded and the next joint configuration is tested. If the path is collision free, the position is accepted as a viable grasp pose.

\subsubsection{Additional sorting of poses} 
When calculating the grasp poses an additional sorting of the poses is performed. This is done to optimize the success rate of the grasping operations without limiting the number of possible solutions. 

The sorting is made to prioritize grasps without any collision. The collision types are sorted into two types, collision with objects and collision with the bin. By allowing for these collisions the space of possible solutions increases, avoiding the situation were an object cannot be grasped at all, and the bin, therefore, cannot be emptied. However, it is expected that the failure rate for such grasps will be much higher. Thus we prioritize collision free grasps, to reduce the overall run-time.

The object collisions are found using the depth data from the sensor. By projecting the fingers into the point cloud data any collision with the objects can be found. Collisions with the bin are found using the workcell model. The method does not accept any collisions with the model, but if by moving the fingers back 2cm in TCP space removes the collision there is a great change that the grasp will succeed. Thus the search for viable grasp poses if first performed with collision free poses $T_{none}$, then poses with collision with objects are used $T_{obj}$, and finally poses with collision with the bin are used $T_{bin}$. The complete set of poses is thus as in Eq.~\ref{eq:posesorting}.

\noindent

\begin{equation} \label{eq:posesorting}
T_{poses} = \{ T_{none} ; T_{obj}; T_{bin} \}
\end{equation}








\subsection{The Grasp Executor} \label{met:exe}

\begin{figure}[ht]
\begin{center}
    \vspace{1.5mm}
   \includegraphics[width=0.65\linewidth]{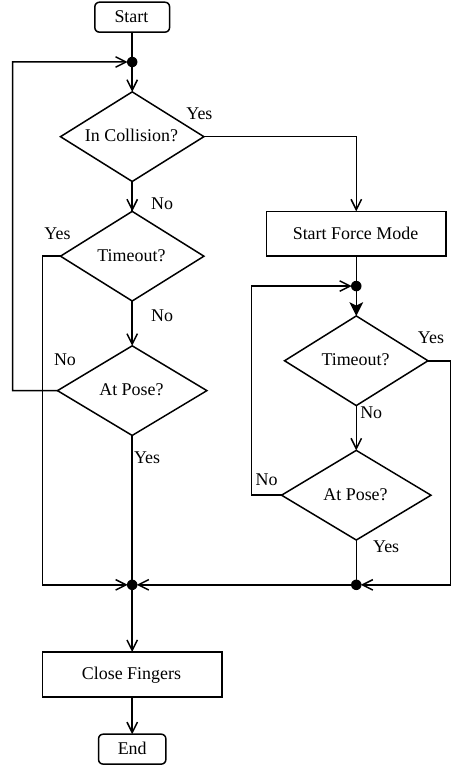}
   \caption{The workflow when a grasp is performed. The ideal grasp moves to the intended pose, and force mode is only activated if collision is detected.}
   \label{fig:grasp_workflow}
     \vspace{-6mm}
\end{center}
\end{figure}

The grasp execution is created to compensate for the fact that collision can occur while grasping the object. Thus the robot is moving to the grasp pose in collision mode. If collision is detected the robot moves into force mode and moves towards the grasp pose. At either timeout, or when reaching the position the robot stops moving. 
The fingers are then closed, and a grasp can be verified by checking that they are not completely closed. 
Before starting the grasp, the fingers are opened according to the object size so as decrease collision with other objects.
The procedure is shown in Fig.~\ref{fig:grasp_workflow}.

\section{Experiments}
\label{exp}

\begin{figure}[ht]
\begin{center}
   \includegraphics[width=0.99\linewidth]{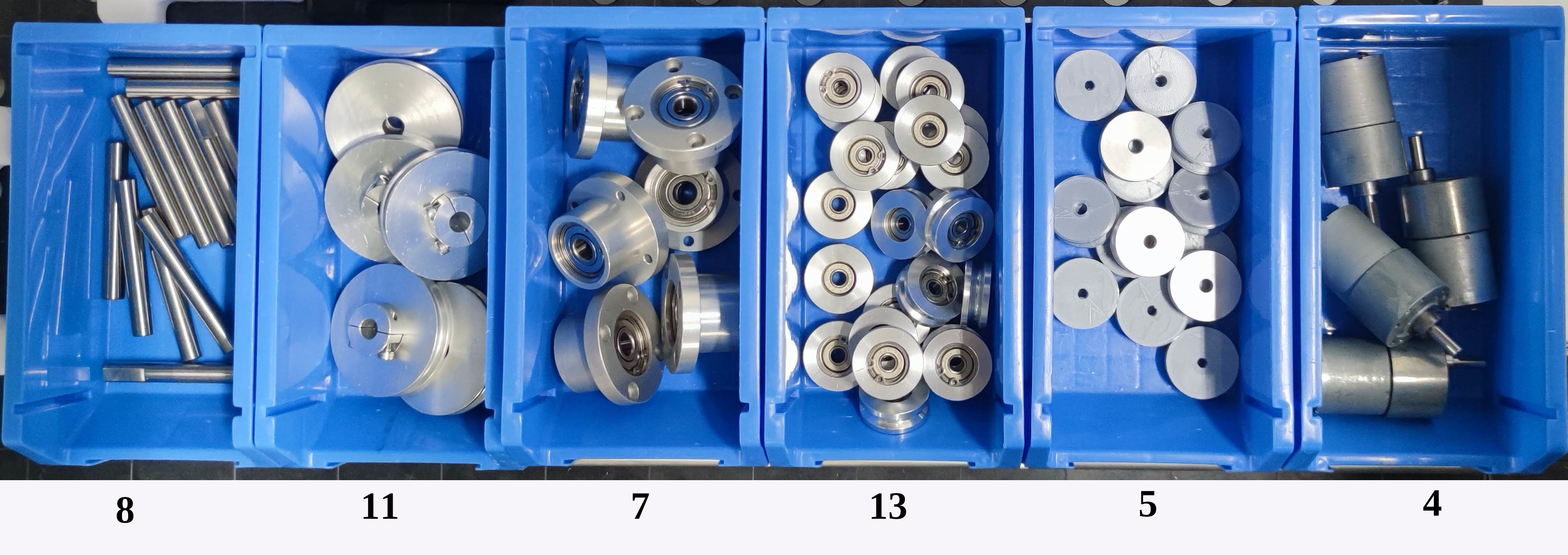}
   \caption{The experimental set-up of the six bins. The ids are according to \cite{yokokohji2019assembly}.}
   \label{fig:objects}
    \vspace{-6mm}
\end{center}
\end{figure}

To demonstrate the performance of the developed workcell, tests are performed on the six large objects from WRSAC18. The number of objects in the bins are set to resemble the lineup at WRSAC18 \cite{yokokohji2019assembly}. The objects in bins are shown in Fig.~\ref{fig:objects}. In the experiments the vision system is used for pose estimation and the grasp planner computes the optimal grasp pose. The grasp executor then finally performs the grasping. 

\subsection{Kitting Task}

To demonstrate that the workcell is able to perform the kitting task, we replicated the requirements of the competition, with three sets of trays to complete. Each set of trays were removed after completion and new empty trays were inserted. During the full run no manual manipulation were performed on the workcell expect for the replacement of the trays.
%
%
The task is run with all large objects, giving 18 object instances to be bin picked. The kitting was performed successfully with 9 failed grasp attempts. However, each failure was recovered from. With two points per object this results in 32 points in all. But, at the competition only 12 large objects were to be bin-picked, thus a maximum of 24 points could be obtained. However, as the best scoring team at the competition obtained 20 points our method still outperforms the other methods.

The full kitting task had a 20 minute run-time, with 30 objects in all, allowing an average per object run-time of 40 seconds. 
The complete run-time of our system for the 18 objects was 8 minutes and 32 seconds. Thus each bin picking lasted 28 seconds on average, well under the 40 seconds limit for each object. This indicates that the remaining points can still be obtained, and our method is appropriate to complete the full task.

\subsection{Analysis of performance}

To analyze the performance of the workcell extensive testing is performed. For each object 
more than one hundred bin-pickings are executed. The bin-picking performed until the bin was emptied, the bin was then refilled, the objects were scrambled, and the bin-picking resumed. 

During the bin-picking we record the overall success rate, and the success rate based on whether no collision was predicted, or if collision was predicted with objects or with the bin. 
We also record the success rate based on whether collision actually occurred, and if the grasping reached time-out. Additionally, we present the systems ability to predict collisions correctly. 
The results of the experiments are shown in Tab.~\ref{tab:wrs}.

\begin{table}[ht]
    \begin{center}
    \caption{
    Analysis of the Bin picking success rate for the different objects. The success is shown overall, when collision is predicted and not, and when actual collision occurred and not, and when the grasp resulted in timeout. Additionally, we show the accuracy of the system for predicting collision. 
    }

    \begin{tabular}{|l|c|c|c|c|c|c|}
        \hline
        Number   & 4   & 11    & 13    & 7    & 8   & 5    \\ 
        Type    & {\scriptsize Motor} & {\scriptsize Pulley} & {\scriptsize Idler} & {\scriptsize Bearing} & {\scriptsize Shaft} & {\scriptsize Pulley} \\ 
        \hline
        \hline
        Overall          & 78.7    & 69.4       & 85.1 & 85.0 & 80.0 & 80.3 \\ \hline
        No Col. Pred.    & 100     & 76.5       & 96.7 & 87.5 & 100 & 98.6 \\ 
        Obj. Col. Pred.    & 81.4     & 57.1       & 50.0 & 79.3 & 100 & 53.8 \\ 
        Bin. Col. Pred.    & 78.8     & 42.95       & 100 & 50.0 & 73.1 & 58.6 \\ \hline
        
        No Collision       & 100     & 77.7       & 97.0 & 90.7 & 85.4  & 96.9 \\ 
        Collision          & 50.0    & 40.6       & 57.1 & 55.0 & 71.4  & 29.0 \\ 
        Timeout            & 27.8    & 20.0       & 40.0 & 14.3 & 62.3  & 16.7 \\ \hline
        \hline
        Pred. Acc          & 77.0    & 79.2       & 85.1 & 77.2 & 72.0 & 76.4 \\ \hline
        \end{tabular}

        \label{tab:wrs}
    \end{center}
\end{table}

From the experiments several interesting observations are made. It is clear that the collision prediction improves the performance. The success rate when no collision is predicted is much higher than when collision is predicted. It is thus valuable to prioritize poses without collision. Additionally, while the success rate decreases when a collision is predicted, the system is often able to perform a grasp. Thus if objects are only placed in positions which will result in collisions the system is still able to perform grasps. This combined with the systems ability to recover from failures allows the system empty bins even if objects are placed in challenging grasp poses.

Additionally, the ability to grasp when actual collisions occurs is shown. While collision free grasps are more successful, the method is still able to grasp objects in collision. The success-rate for grasps when the timeout was reached is even lower, but grasps still succeed. Increasing the timeout run-time could increase the success-rate but increase the overall run-time. This is an interesting topic for further research.

The accuracy of actually predicting a collision is between 72.0~\% and 85.1~\%. This shows that while improving performance the collision prediction is also fairly accurate.





\subsubsection{Grasp Types}

\begin{figure}[ht]
    \vspace{1.5mm}
    \begin{center}
        \hfill
        \begin{subfigure}[t]{.16\textwidth}
          \centering
          \includegraphics[trim=0 0 0 0,clip,width=0.99\linewidth]{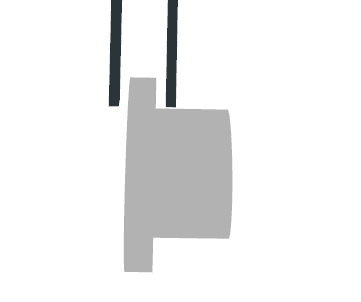}
          \caption{Type 1}
          \label{fig:type:01}
        \end{subfigure}%
        \hfill
        \begin{subfigure}[t]{.16\textwidth}
          \centering
          \includegraphics[trim=40 40 40 40,clip,width=0.99\linewidth]{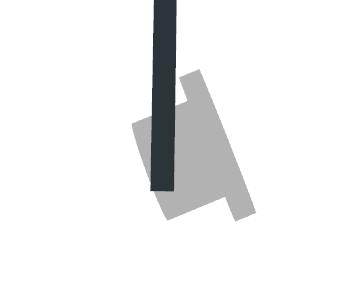}
          \caption{Type 2}
          \label{fig:type:02}
        \end{subfigure}%
        \hfill
        \begin{subfigure}[t]{.16\textwidth}
          \centering
          \includegraphics[trim=0 0 0 0,clip,width=0.99\linewidth]{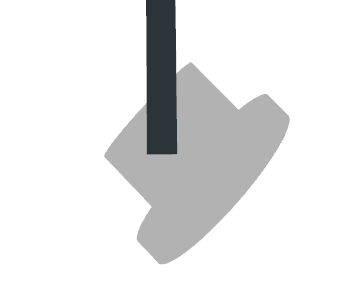}
          \caption{Type 3}
          \label{fig:type:03}
        \end{subfigure}%
        \hfill
    
        \hfill
        \begin{subfigure}[t]{.16\textwidth}
          \centering
          \includegraphics[trim=0 0 0 0,clip,width=0.99\linewidth]{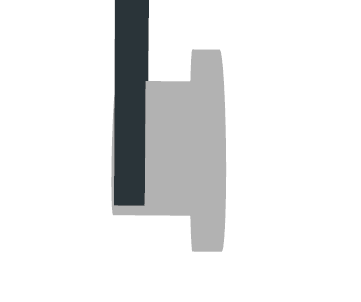}
          \caption{Type 4}
          \label{fig:type:04}
        \end{subfigure}%
        \hfill
        \begin{subfigure}[t]{.16\textwidth}
          \centering
          \includegraphics[trim=0 0 0 0,clip,width=0.99\linewidth]{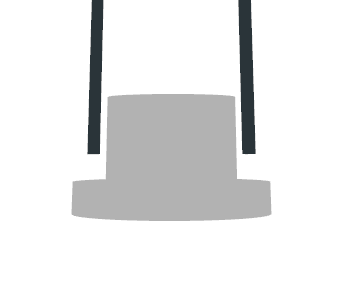}
          \caption{Type 5}
          \label{fig:type:05}
        \end{subfigure}%
        \hfill
        \begin{subfigure}[t]{.16\textwidth}
          \centering
          \includegraphics[trim=0 0 0 0,clip,width=0.99\linewidth]{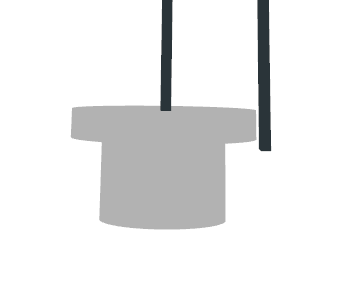}
          \caption{Type 6}
          \label{fig:type:06}
        \end{subfigure}%
        \hfill
       
       \caption{The six different grasps defined for object 7. Type 2, 3, 4 and 5 are similar with the relative angle to z varying.}
       \label{fig:type}
         \vspace{-6mm}
    \end{center}
\end{figure}

\begin{table}
\begin{center}
\caption{
    Results for the different grasp types of object 7.
    }
    \begin{tabular}{|l|c|c|c|c|c|c|}
    \hline
    Type   & 1 & 2 & 3 & 4 & 5 & 6 \\ \hline
    \hline
    Attempts & 51 & 11 & 3 & 7 & 17 & 38 \\ \hline
    Success & 50 & 10 & 2 & 4 & 16 & 24 \\ \hline
    Collision & 2 & 0 & 1 & 0 & 3 & 14 \\ \hline
     \end{tabular}
    \label{tab:type}
    \end{center}
 \vspace{-4mm}
\end{table}

To further analyze the results of grasps, the results for object 7 are split into grasp types. For the object six different grasps have been created, the grasps are shown in Fig.~\ref{fig:type}. 
%
%
The results for each of the different grasp types are shown in Tab.~\ref{tab:type}. From the results several interesting observations are noted. All grasps are used during the process so having, a full coverage is important.
Type 1 is the most used grasp, which indicates that the object often ends up in a position where this is the most desirable grasp. This is also seen by the fact that the number of collisions is very low. The second most applied grasp type is 6. However, here the success-rate is lower, and the collision rate is higher. This is possibly as a result of collision with the object when trying to insert the finger.  
Grasps 2 trough 5 are all variations of the same grasp with 2 and 5 being the most prevalent. These are possibly the most stable poses in the bin.




\section{Conclusion}
\label{con}

This paper presented a workcell for off-the-shelf bin-picking. The bin-picking abilities are tested on the large objects from the World Robot Summit Assembly Challenge 2018. Here the workcell show state-of-the-art performance, by successfully kitting of all objects within the timeline.

To accomplish this task a colorless pose estimation approach using point clouds has been developed. Combined with a grasp estimation planner grasp poses are obtained. Our developed grasp executor then completes the grasp and adapt to any collisions. Extensive testing shows the validity of our approach.
We believe that these results demonstrate the progress of robotics and computer vision in recent years. 

%
%


In further work, it would be interesting to apply zero-shot pose estimation methods for the workcell. This could be combined with using real training data. While manual data collection is very expensive this could be performed automatically. As the bin-picking successfully grasps objects the feedback could be used to collect data and improve the pose estimation algorithm. 





Another interesting work would be the completion of the full kitting challenge. However, the smaller objects might introduce a need for more components such as re-grasping or fixtures. 
Additionally, novel large objects could be introduced and the workcell performance could be tested.

\bibliographystyle{IEEEtran} 
\bibliography{egbib}

\end{document}